\documentclass{article}

    \usepackage[preprint]{neurips_2022}

\usepackage[utf8]{inputenc} 
\usepackage[T1]{fontenc}    
\usepackage{hyperref}       
\usepackage{url}            
\usepackage{booktabs}       
\usepackage{amsfonts}       
\usepackage{nicefrac}       
\usepackage{microtype}      
\usepackage{xcolor}         

\usepackage{multirow}
\usepackage{algorithm}
\usepackage{algorithmic}
\usepackage{multirow}
\usepackage{tabularx}
\usepackage{amsfonts,bm}
\usepackage{threeparttable}
\usepackage{xspace}
\usepackage{amsmath}
\usepackage{upgreek}
\usepackage{amssymb}
\usepackage{pifont}%
\usepackage{multirow}
\usepackage{makecell}
\usepackage{color}
\usepackage{graphicx}
\usepackage{colortbl}
\usepackage{wrapfig}
\usepackage{soul}
\newcommand{\tabincell}[2]{\begin{tabular}{@{}#1@{}}#2\end{tabular}}  
\newcommand{\xmark}{\ding{55}}%

\newlength\savewidth

\title{HiViT: Hierarchical Vision Transformer Meets Masked Image Modeling}

\author{%
  Xiaosong Zhang$^{1}$\thanks{Equal Contribution.}
  \And
  Yunjie Tian$^{1}$\footnotemark[1]
  \And
  Wei Huang$^{1}$
  \And
  Qixiang Ye$^{1}$
  \And
  Qi Dai$^{2}$
  \AND
  Lingxi Xie$^{3}$
  \And
  Qi Tian$^{3}$
  \AND \\ University of Chinese Academy of Sciences$^1$
  \And \\ Fudan University$^2$
  \And \\ Huawei Inc.$^3$
}

\begin{document}

\maketitle

\begin{abstract}
Recently, masked image modeling (MIM) has offered a new methodology of self-supervised pre-training of vision transformers. A key idea of efficient implementation is to discard the masked image patches (or tokens) throughout the target network (encoder), which requires the encoder to be a plain vision transformer (\textit{e.g.}, ViT), albeit hierarchical vision transformers (\textit{e.g.}, Swin Transformer) have potentially better properties in formulating vision inputs. 
{\color{black}
In this paper, we offer a new design of hierarchical vision transformers named \textbf{HiViT} (short for Hierarchical ViT) that enjoys both high efficiency and good performance in MIM. 
The key is to remove the unnecessary `local inter-unit operations', deriving structurally simple hierarchical vision transformers in which mask-units can be serialized like plain vision transformers. 
}
%
Empirical studies demonstrate the advantageous performance of HiViT in terms of fully-supervised, self-supervised, and transfer learning. 
In particular, in running MAE on ImageNet-1K, HiViT-B reports \textbf{a $\mathbf{+0.6\%}$ accuracy gain} over ViT-B and \textbf{a $\mathbf{1.9\times}$ speed-up} over Swin-B, and the performance gain generalizes to downstream tasks of detection and segmentation.
Code will be made publicly available.
\end{abstract}

\section{Introduction}
\label{intro}

Deep neural networks have been the fundamentals of deep learning~\cite{lecun2015deep} and advanced the research fields of computer vision, natural language processing, \textit{etc.}, in the past decade. Recently, the computer vision community has witnessed the emerge of vision transformers~\cite{ViT2021,Swin2021,wang2021pvtv2,zhou2021deepvit,dai2021coatnet,li2021efficient}, transplanted from the language models~\cite{Attention2017,devlin2019bert}, that replaced the dominance of convolutional neural networks~\cite{AlexNet2012,Resnet2016,EfficientNet2019}. They have the ability of formulating long-range feature dependencies, which naturally benefits visual recognition especially when long-range relationship is important.

There are mainly two families of vision transformers, namely, the plain vision transformers~\cite{ViT2021,DeiT2021} and the hierarchical vision transformers~\cite{Swin2021,wang2021pvtv2,dong2021cswin, chen2021crossvit}, differing from each other in whether multi-resolution feature maps are used.
{\color{black}
While the latter is believed to capture the nature of vision signals (most convolution-based models have used the hierarchical configuration), but used some spatial local operations (\textit{i.e.}, early-stage self-attentions with shifting window). These models can encounter difficulties when the tokens need to be flexibly manipulated. A typical example lies in masked image modeling (MIM), a recent methodology of pre-training vision transformers~\cite{bao2021beit,MAE2021,xie2021simmim} -- a random part of image patches are hidden from input, and it is difficult for the hierarchical models to determine whether each pair of tokens need to communicate, unlike the plain models. 
Essentially, this is because hierarchical vision transformers have used non-global operations (\textit{e.g.}, window attentions) between the masking units\footnote{\textcolor{black}{The minimum size of the masked pixels when executing MIM is defined as masking unit. For example, when the input image size is $224\times224$, the masking unit size for MAE~\cite{MAE2021} is $16\times16$.}}. Hence, unlike the plain vision transformers that can serialize all tokens for acceleration, the hierarchical vision transformers must maintain the two-dimensional structure, keeping the dummy (masked) tokens throughout the encoder. Consequently, as shown in~\cite{xie2021simmim}, the training speed of hierarchical transformers is $2\times$ slower than that using plain transformers, and very few works chose to follow this direction.
}

%
%
%
{\color{black}
In this paper, we start with categorizing the operations in hierarchical vision transformers into `intra-unit operations', `global inter-unit operations', and `local inter-unit operations'.
We note that plain vision transformers only contain `intra-unit operations' (\textit{i.e.}, patch embedding, layer normalization, MLP) and `global intra-unit operations' (\textit{i.e.}, global self-attentions), hence the units' spatial coordinates can be discarded and the units can be serialized for efficient computation, like in MAE~\cite{MAE2021}. 
That said, for hierarchical vision transformers, it is the `local inter-unit operations' (\textit{i.e.}, shifting-window self-attentions, patch merging) that calls for extra judgment based on the units' spatial coordinates and obstructs the serialization as well as removing the masked units.

A key observation of this paper lies in that `local inter-unit operations' do not contribute much for recognition performance -- what really makes sense is the hierarchical design (\textit{i.e.}, multi-scale feature maps) itself. Hence, to fit hierarchical vision transformers to MIM, we remove the `local inter-unit operations', resulting in a simple hierarchical vision transformer that absorbs both the flexibility of ViT~\cite{ViT2021} and the superiority of Swin Transformer~\cite{Swin2021}. 
There are usually 4 stages of different resolutions in hierarchical vision transformers where the 3rd stage has the largest number of layers and we call it the main stage. 
{\color{black}
We remove the last stage of Swin and switch off all local inter-unit window attentions, only keeping the global attention between tokens in the main stage.}
In practice, the last stage is merged into the main stage (to keep the model FLOPs unchanged) and local window attentions in the early-stage are replaced by an intra-unit multi-layer perceptron with same FLOPs. 
With these minimal modifications, we remove all redundant `local inter-unit operations' in hierarchical vision transformers, where only the simplest hierarchical structure is adopted. 
Compared to the plain ViTs, our model only adds only several spatial merge operations and MLP layers before the main stage. 
The resulting architecture is named \textbf{HiViT} (short for Hierarchical ViT), which has the ability of modeling hierarchical visual signals yet all tokens are maximally individual and remain flexibility for manipulation. 
Meanwhile, our HiViT maintains the ViT paradigm, which is very simple to implement compared to other hierarchical vision transformers. 
}

\begin{wrapfigure}{r}{0.45\linewidth}
    \vspace{-0.4cm}
    \centering
    \includegraphics[
    width=1.0\linewidth, trim=0 1 0 1, clip
    ]{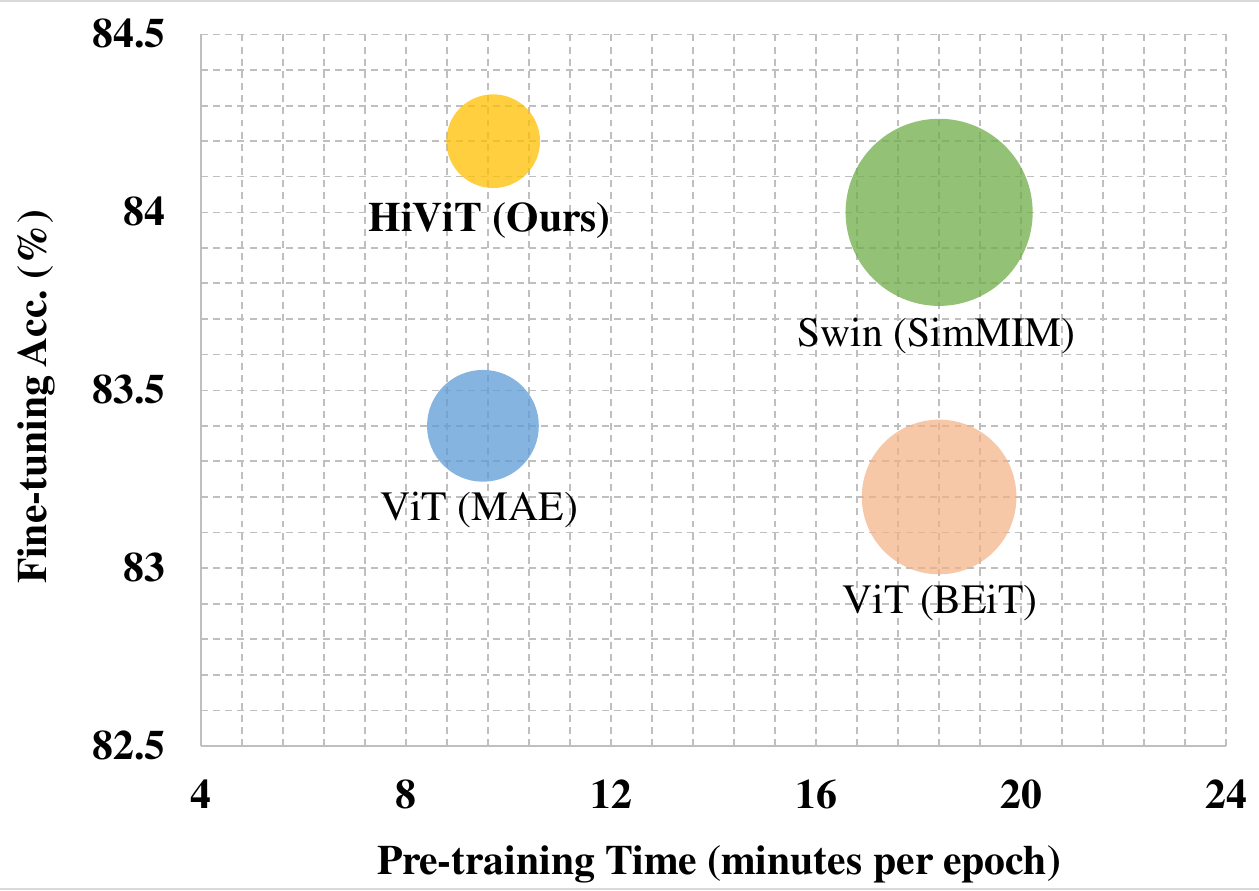}
    \caption{
    Self-supervised pre-training of HiViT is significantly faster than Swin with SimMIM~\cite{xie2021simmim} and the result is better than ViT trained with MAE \cite{MAE2021} and BEiT~\cite{bao2021beit}.   Circle size denotes memory requirement. All the models are in base scale.
    }
    \label{fig:efficiency}
    \vspace{-0.4cm}
\end{wrapfigure}

We perform fully-supervised classification experiments on ImageNet-1K to validate the superiority of HiViT. Lying between ViT and Swin Transformer, HiViT enjoys consistent accuracy gains over both the competitors, \textit{e.g.}, HiViT-B reports a $83.8\%$ top-1 accuracy, which is $+2.0\%$ over ViT-B and $+0.3\%$ over Swin-B. 
%
%
{\color{black}
With extensive ablation studies, we find that removing the `local inter-unit operations' does not harm the recognition performance, yet the hierarchical structure and relative positional encoding (not `local inter-unit operations') slightly but consistently improves the classification accuracy. This makes HiViT applicable to a wide range of visual recognition scenarios. 
}

Continuing to MIM, the advantages of HiViT become clearer. 
With $800$ epochs of MIM-based pre-training and $100$ epochs of fine-tuning, HiViT-B reports $84.2\%$ top-1 accuracy on ImageNet-1K, {\color{black}which is $+0.6\%$ over ViT-B (using MAE~\cite{MAE2021}, pre-training for $1600$ epochs) and $+0.2\%$ over Swin-B (using SimMIM~\cite{xie2021simmim}). More importantly, HiViT enjoys the efficient implementation that discards {\color{black}all masked patches (or tokens) at the input stage}, and hence the training speed is $1.9\times$ as fast as that of SimMIM, since the original Swin Transformer must forward-propagate the full token set (Fig.~\ref{fig:efficiency}).} The advantages persist to other visual recognition tasks, including linear probing ($71.3\%$ top-1 accuracy) on ImageNet-1K, semantic segmentation ($48.3\%$ mIoU) on the ADE20K dataset~\cite{zhou2017scene}, and object detection ($49.5\%$ AP) and instance segmentation ($43.8\%$ AP) on the COCO dataset~\cite{lin2014microsoft} ($1\times$ training schedule). These results validate that removing `local inter-unit operations' does not harm generic visual recognition.

\textcolor{black}{The core contribution of this paper is HiViT, a hierarchical vision transformer architecture that is off-the-shelf for a wide range of vision applications. In particular, with masked image modeling being a popular self-supervised learning paradigm, HiViT has the potential of being directly plugged into many existing algorithms to improve their effectiveness and efficiency in learning visual representations from large-scale, unlabeled data.}

\section{Related Work}
\label{related}

\subsection{Vision Transformers}

Vision transformers~\cite{ViT2021} were adapted from the natural language processing (NLP) transformers~\cite{Attention2017,devlin2019bert}, opening a new direction of designing visual recognition models with weak induction bias~\cite{mlp-mixer}. Early vision transformers~\cite{ViT2021} mainly adopted the plain configuration, and efficient training methods are strongly required~\cite{DeiT2021}. To cater for vision-friendly priors, Swin Transformer~\cite{Swin2021} proposed a hierarchical architecture that contains multi-level feature maps and validated good performance in many vision problems. Since then, various efforts emerged in improving hierarchical vision transformers, including borrowing design experiences from convolutional neural networks~\cite{wang2021pvtv2,wu2021cvt,vaswani2021scaling}, adjusting the design of self-attention geometry~\cite{dong2021cswin, yang2021focal}, designing hybrid architectures to integrate convolution and transformer modules~\cite{srinivas2021bottleneck,container,dai2021coatnet,Conformer2021}, \textit{etc}.

Essentially, there is a tradeoff between plain and hierarchical vision transformers -- in terms of whether strong induction bias is to be introduced. As we shall see later, the increase of induction bias may weaken the flexibility and thus efficiency of applying vision transformers to particular scenarios (\textit{e.g.}, masked image modeling). In this paper, we design a hierarchical vision transformer that maximally discards induction bias, achieving both high efficiency and good performance.

\subsection{Self-Supervised Learning and Masked Image Modeling}

In the context of computer vision, self-supervised learning aims to learn compact visual representations from unlabeled data. The key to this goal is to design a pretext task that sets a natural constraint for the target model to achieve by tuning its weights. The existing pretext tasks are roughly partitioned into three categories, namely, \textbf{geometry-based} proxies that were built upon the spatial relationship of image contents~\cite{wei2019iterative,jigsaw,rotation}, \textbf{contrast-based} proxies that assumed that different views of an image shall produce related visual features~\cite{he2020momentum,chen2020simple,grill2020bootstrap, caron2021emerging,swav,pixpro,sage}, and \textbf{generation-based} proxies that required visual representations to be capable of recovering the original image contents~\cite{colorization,inpainting,MAE2021,bao2021beit,beyond}. After the self-supervised learning (\textit{a.k.a.} pre-training) stage, the target model is often evaluated by fine-tuning in a few downstream recognition tasks -- the popular examples include image classification at ImageNet-1K~\cite{ImageNet2009}, semantic segmentation at ADE20K~\cite{zhou2017scene}, object detection and instance segmentation at COCO~\cite{COCO2014}, \textit{etc}.

We are interested in a particular generation-based method named masked image modeling (MIM)~\cite{bao2021beit,MAE2021}. The flowchart is straightforward: some image patches (corresponding to tokens) are discarded, the target model receives the incomplete input and the goal is to recover the original image contents. MIM is strongly related to the masked language modeling (MLM) task in NLP. BEiT~\cite{bao2021beit} transferred the task to the computer vision community by masking the image patches and recovering the tokens produced by a pre-trained model (knwon as the tokenizer). MAE~\cite{MAE2021} improved the MIM framework by only taking the visible tokens as input and computing loss at the pixel level -- the former change largely accelerated the training procedure as the computational costs of the encoder went down. The follow-up works explored different recovery targets~\cite{wei2021masked}, more complicated model designs~\cite{CIM,chen2020simple}, and other pretext tasks.

{\color{black}It is worth noting that MIM matches plain vision transformers very well because each token is an individual unit and only the unmasked tokens are necessary during the pre-training process.} The properties does not hold for hierarchical vision transformers, making them difficult to inherit the good properties (\textit{e.g.}, training efficiency). Although SimMIM~\cite{xie2021simmim} tried to combine Swin Transformer with MIM, it {\color{black}uses all tokens}, including those corresponding to the masked patches, shall be preserved during the encoder stage, incurring much heavier computational costs. In this paper, we present a hierarchical vision transformer that is free of such burden.

\section{Hierarchical Vision Transformer for Masked Image Modeling}
\label{method}

\subsection{Preliminaries}

Masked image modeling (MIM) is an emerging paradigm of self-supervised visual representation learning. The flowchart involves feeding a partially masked image to the target model and training the model to recover it. Mathematically, let the target model be $f\!\left(\mathbf{x};\boldsymbol{\theta}\right)$ where $\boldsymbol{\theta}$ denotes the learnable parameters. Given a training image, $\mathbf{x}$, it is first partitioned into a few patches, $\mathcal{X}=\left\{\mathbf{x}_1,\mathbf{x}_2,\ldots,\mathbf{x}_M\right\}$, where $M$ is the number of patches. Then, MIM randomly chooses a subset $\mathcal{M}'\subset\left\{1,2,\ldots,M\right\}$, feeds the patches with IDs in $\mathcal{M}'$ (denoted as $\mathcal{X}'$) into the target model $f\!\left(\mathbf{x};\boldsymbol{\theta}\right)$ (\textit{a.k.a.}, the encoder), and appends a decoder to it, aiming at recovering the original image contents, either tokenized features~\cite{bao2021beit} or pixels~\cite{MAE2021}, at the end of the decoder. If $f\!\left(\mathbf{x};\boldsymbol{\theta}\right)$ is able to solve the problem, it is believed that the parameters have been well trained to extract compact visual features.

An efficient vision model that fits MIM is the vanilla vision transformer, abbreviated as ViT~\cite{ViT2021}. In ViT, each image patch is transferred into a token (\textit{i.e.}, a feature vector), and the tokens are propagated through a few transformer blocks for visual feature extraction. Let there be $L$ blocks, the $l$-th block takes the token set of $\mathcal{U}^{(l-1)}$ as input and outputs $\mathcal{U}^{(l)}$, and $\mathcal{U}^{(0)}\equiv\mathcal{X}$. The main part of each block is self-attention, for which three intermediate features are computed upon $\mathbf{u}_m^{(l-1)}$, namely the query, key, and value, denoted as $\mathbf{q}_m^{(l-1)}$, $\mathbf{k}_m^{(l-1)}$, and $\mathbf{v}_m^{(l-1)}$, respectively. Based on these quantities, the self-attention of $\mathbf{z}_m^{(l-1)}$ is computed by $\mathrm{SA}\!\left(\mathbf{z}_m^{(l-1)}\right)=\mathrm{softmax}\!\left[\mathbf{q}_m^{(l-1)}\cdot\mathbf{k}_1^{(l-1)\top},\ldots,\mathbf{q}_m^{(l-1)}\cdot\mathbf{k}_M^{(l-1)\top}\right]/\sqrt{D_\mathrm{key}}\cdot\left[\mathbf{v}_1^{(l-1)},\ldots,\mathbf{v}_M^{(l-1)}\right]^\top$, where $1/\sqrt{D_\mathrm{key}}$ is a scaling vector. Auxiliary operations, including layer normalization, multi-layer perceptron, skip-layer connection, are applied after the self-attention computation. ViT has been applied to a series of vision problems, but we emphasize its particular efficiency on MIM, which lies in that the tokens not in $\mathcal{X}'$ can be discarded at the beginning of encoder, decreasing the complexities of the pre-training process by a factor of $M/\left|\mathcal{M}'\right|$ (\textit{e.g.}, $4$ in the regular setting of MAE~\cite{MAE2021}).

Intuitively, hierarchical vision transformers (\textit{e.g.}, Swin Transformer) are better at capturing multi-level visual features. 
It has three major differences from ViT: 
(i) the architecture is partitioned into a few stages and the spatial resolution, rather than being fixed, is gradually shrunk throughout the forward propagation; 
(ii) to handle relatively large token maps, the self-attention computation is constrained within a grid of windows, and {\color{black}the window partition is shifted across layers;} 
(iii) global positional encoding is replaced by relative positional encoding -- this is to fit the window attention mechanism. 
Although hierarchical vision transformers report higher visual recognition accuracy, these models are not so efficient as ViT in terms of MIM, and the reasons are revealed in the next part. Consequently, few prior works have tried the combination -- as an example, SimMIM~\cite{xie2021simmim} fed the entire image (the masked patches are replaced with mask tokens which are learnable) into the encoder, resulting in heavier computational costs in time and memory.


\begin{figure}[!t]
    \centering
    \includegraphics[width=1.0\linewidth]{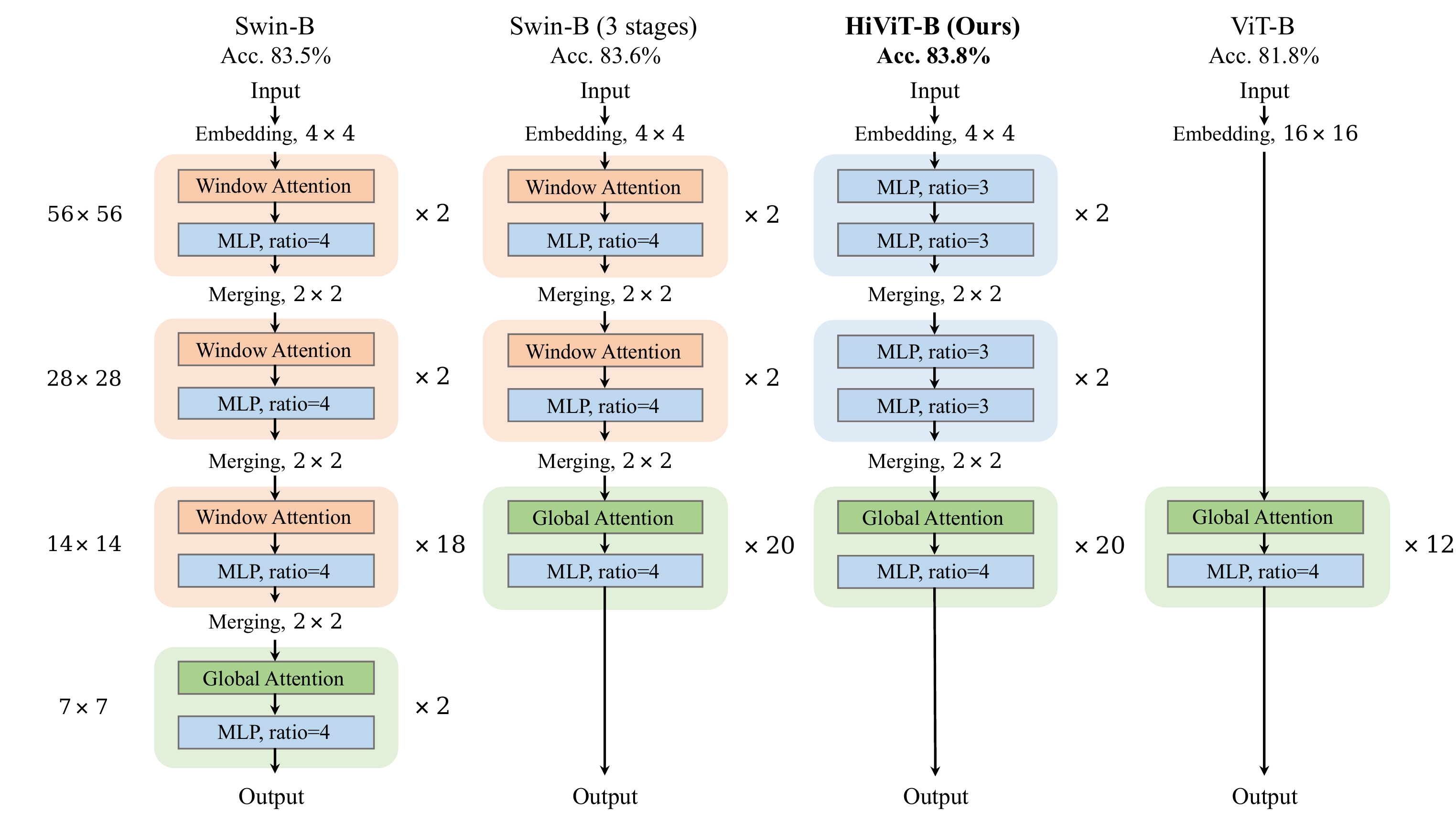}
    \caption{Comparison of the architectures of Swin Transformer, ViT, and the proposed HiViT.}
    \label{fig:hivit}
\end{figure}

\subsection{HiViT: Efficient Hierarchical Transformer for MIM}

{\color{black}

We pursue for the efficient implementation of MAE~\cite{MAE2021}, \textit{i.e.}, only the active (unmasked) tokens are fed into the encoder -- mathematically, we are always dealing with a squeezed list of $\left|\mathcal{M}'\right|$ tokens. 
The major difficulty of integrating it with hierarchical vision transformers (\textit{e.g.}, Swin Transformers) lies in the `local inter-unit operations', which make it difficult to serialize the tokens and abandon the inactive (masked) ones. To remove them, we first set the masking unit size to be the token size at the main stage -- for Swin Transformers, this is the 3rd stage that is the major part (\textit{e.g.}, for Swin-B, the 3rd stage has $18$ blocks, and the entire architecture has $24$ blocks). {\color{black}The masking unit is $16\times16$ pixels that aligns with the constant token size of ViT.} Then, we adjust the model as follows:
}
\begin{itemize}
\item For the operations after the main stage, we do not allow the patch merging that mixes active and inactive patches. For simplicity, we directly remove the last (4th) stage in the Swin Transformer, which has only two blocks, and append the same number of blocks to the 3rd stage. Since the 3rd stage has a smaller token dimensionality, such an operation saves both trainable parameters, while changing the `local inter-unit operations' in the 3rd stage to `global inter-unit operations'. As a result, the model's recognition performance becomes better (Fig.~\ref{fig:hivit}).
\item For the operations prior to the main stage, we do not allow the window attention in the former 2 stages. That said, we remove the shift window of Swin and do not introduce any other `local inter-unit operations' such as window attentions or convolutions. As an alternative, we only employ MLP block (replacing the self-attention by another mlp layer) for the former 2 stages. Surprisingly, as demonstrated in Fig.~\ref{fig:hivit}, this modification brings $0.2\%$ performance improvement without bells and whistles. Compared to plain ViT as shown in Fig.~\ref{fig:hivit}, the derived architecture possesses hierarchical property and only requires two MLP blocks but enjoys much better performance on both self-/fully-supervised learning. 
\end{itemize}


The above procedure produces an architecture between Swin Transformer (hierarchical) and ViT (plain). We illustrate the procedure in Figure~\ref{fig:hivit}. The resulting architecture, named HiViT (short for Hierarchical ViT), \textbf{is structurally simple and brings an efficient implementation for MIM.} Specifically, HiViT abandons all the `local inter-unit operations' in the entire architecture, therefore {\color{black} the masked image patches can be discarded at the input layer and its computation can be eliminated in all stages.
}
As a result, HiViT enjoys both the effectiveness of hierarchical vision transformers to capture visual representations (\textit{i.e.}, the recognition accuracy is much higher than ViT) and the efficiency of plain vision transformers in the masked image modeling task (\textit{i.e.}, the efficient implementation of MAE~\cite{MAE2021} can be directly transplanted, making HiViT almost $2\times$ faster than Swin Transformer in MIM). Detailed results are provided in the experimental part.


\begin{table}[t]
    \caption{Configurations for HiViT variants.}
    \centering
    \vspace{-4pt}
    \label{tab.config}
    \begin{tabular}{l|ccc|ccc|ccc|cc}
    \toprule
    \multirow{2}{*}{Model}  & \multicolumn{3}{c|}{Depth}     & \multicolumn{3}{c|}{Dim}      & \multicolumn{3}{c|}{Heads}      & Params  & FLOPs   \\
                            & $56^2$   & $28^2$   & $14^2$   & $56^2$   & $28^2$   & $14^2$  & $56^2$   & $28^2$   & $14^2$  & (M)     & (G)     \\
    \midrule
    HiViT-T (Tiny)                 & 1        & 1        & 10       & 96       & 192      & 384     & -        & -        & 6       & 19.2    & 4.6     \\
    HiViT-S (Small)                 & 2        & 2        & 20       & 96       & 192      & 384     & -        & -        & 6       & 37.5    & 9.1     \\
    HiViT-B (Base)                 & 2        & 2        & 20       & 128      & 256      & 512     & -        & -        & 8       & 66.4    & 15.9    \\
    \bottomrule
\end{tabular}
\end{table}

\section{Experiments}
\label{experiment}

We first conduct fully supervised experiments with labels using the proposed HiViT on ImageNet dataset~\cite{ImageNet2009}. Then, HiViT models are tested using masked image modeling self-supervised methods (MIM)~\cite{MAE2021}. The self-supervised pre-trained models are also transferred to downstream tasks including object detection on COCO dataset~\cite{COCO2014} and semantic segmentation on ADE20K~\cite{zhou2017scene}. We provide the ablation studies about our methods on both fully-/self-supervised learning.

\subsection{ImageNet Classification with Labels}

\paragraph{Training Settings} We first evaluate our models with fully supervised learning on ImageNet-1K~\cite{ImageNet2009} which contains 1.28M training images and 50K validation ones divided into 1,000 categories. We follow Swin~\cite{Swin2021} and use the same training settings without other tricks. Specifically, we use AdamW optimizer~\cite{adamw} with an initial learning rate of 0.001, a weight decay of 0.05, batch size of 1024, a cosine decay learning rate scheduler, and a linearly warm-up for 20 epochs. All the models are trained for 300 epochs with augmentation and regularization strategies of ~\cite{Swin2021} and exponential moving average (EMA) technique. The input size is $224\times224$ in default. The output feature of 3rd stage is followed by an average pooling layer and then a classifier layer. We adopt drop path rate of $0.05$, $0.3$, and $0.5$ for HiViT-T/S/B.

\begin{table}[t]
\small
\centering
\setlength{\tabcolsep}{1pt}
\renewcommand\arraystretch{1.0}
\caption{Comparison of image classification on ImageNet-1K for different models. 
All models are trained and evaluated with $224 \times 224$ resolution on ImageNet-1K in default, unless otherwise noted. 
}
\vspace{-4pt}
\resizebox{0.42\linewidth}{!}{
    \begin{tabular}{l|ccc}
    \toprule
    \multirow{2}{*}{Model} & Params & FLOPs & Top-1 \\
    & (M) & (G) & (\%) \\
    \midrule
    DeiT-S/16~\cite{DeiT2021} & 22.1 & 4.5 & 79.8 \\
    PVT-S\cite{wang2021pyramid}  & 24.5 & 3.8 & 79.8 \\
    Swin-T~\cite{liu2021swin}   & 28.3 & 4.5 & 81.2 \\
    CvT-13~\cite{wu2021cvt}        & 20.0 & 4.5 & 81.6 \\
    CaiT-XS-24~\cite{touvron2021going} & 26.6 & 5.4 & 81.8 \\
    \midrule
    HiViT-T (Ours)       & 19.2 & 4.6 &\textbf{82.1} \\
    \midrule
    CvT-21~\cite{wu2021cvt}  & 32.0 & 7.1 & 82.5 \\
    UFO-ViT-M~\cite{song2021ufo}  & 37.0 & 7.0 & 82.8 \\
    Swin-S~\cite{liu2021swin}    &  49.6 & 8.7 & 83.1 \\
    ViL-M~\cite{zhang2021multi}  & 39.7 & 9.1 & 83.3 \\
    CaiT-S36~\cite{touvron2021going} & 68.0 & 13.9 & 83.3 \\
    \midrule
    HiViT-S  (Ours) & 37.5 & 9.1 &\textbf{83.5} \\
    \bottomrule
    \end{tabular}
    }
\resizebox{0.44\linewidth}{!}{
    \begin{tabular}{l|ccc}
    \toprule
    \multirow{2}{*}{Model} & Params & FLOPs & Top-1 \\
    & (M) & (G) & (\%) \\
    \midrule
    ResNet-152~\cite{Resnet2016}  & 60.0 & 11.0 & 78.3 \\
    PVT-L~\cite{wang2021pvtv2} & 61.4 & 9.8 & 81.7 \\
    DeiT-B/16~\cite{DeiT2021}& 86.7 & 17.4 & 81.8 \\
    CrossViT-B~\cite{chen2021crossvit} & 104.7 & 21.2 & 82.2 \\
    T2T-ViT-24~\cite{yuan2021tokens}  & 64.1 & 14.1 & 82.3 \\
    CPVT-B~\cite{chu2021conditional}  & 88.0 & 17.6 & 82.3 \\
    TNT-B~\cite{han2021transformer} & 65.6 & 14.1 & 82.8\\
    ViL-B~\cite{zhang2021multi} & 55.7 & 13.4 & 83.2 \\
    UFO-ViT-B~\cite{song2021ufo} & 64.0 & 11.9 & 83.3 \\
    CaiT-M24~\cite{touvron2021going} & 185.9 & 36.0 & 83.4 \\
    Swin-B~\cite{liu2021swin}   & 87.8 & 15.4 & 83.5 \\
    \midrule
    HiViT-B (Ours) & 66.4 & 15.9 & \textbf{83.8} \\
    \bottomrule
    \end{tabular}
    }
    \label{tab:image_classification}
    \vspace{-8pt}
\end{table}


\paragraph{Model Configurations} Three models including HiViT-T/S/B are tested on fully supervised learning and their configurations are shown in Tab.~\ref{tab.config}. The ``Depth'' represents the block number on different stages ($56^2$, $28^2$, and $14^2$ are the 1st, 2nd, and 3rd stage respectively). The ``Dim'' and ``Heads'' represent the dimension and attention head of 3 stages. We align our models on FLOPs, and the parameters are less compared to other works. We report the inference throughput speed in Tab.~\ref{tab:inference} by testing the $224^2$ images on V100 GPU with the same script. 

\vspace{-4pt}
\paragraph{ImageNet Results} The fully supervised training results are shown in Tab.~\ref{tab:image_classification}. Compared to vanilla ViT models, all the HiViT models report dominant results. HiViT-T/B surpass DeiT-S/B models by $2.3\%$ and $2.0\%$ respectively with similar flops and fewer parameters. Compared to its follow-ups, our models still show competitive results. In particular, HiViT-T/S/B beat Swin-T/S/B by $0.9\%$, $0.4\%$, and $0.3\%$ respectively with similar complexities and fewer parameters. All the results do not introduce any other tricks compared to Swin~\cite{Swin2021}. In addition, all our models are parameter friendly. For example, compared to Swin-T/S/B models,  HiViT-T/S/B enjoy $32.2\%$, $24.4\%$ and $24.4\%$ fewer parameters. We note that our models are structurally simple which offer a promising baseline for future research.



\vspace{-7pt}
\paragraph{Ablation Studies}
We conduct fully supervised ablation studies to show the advantages of our method. In this part, we inherit the training settings above and the results are shown in Tab.~\ref{tab.evolve}. The `Setting' represents that the modules we remove from top to bottom. The `Dim' represents the dimension on $14^2$ resolution. The `Depth' represents the block numbers of the corresponding stages. The `RPE' is the relative position embedding and the ``Win. Att.'' denotes whether there are window attentions in the model.
As we can see in the Tab.~\ref{tab.evolve}, removing the last stage (\st{Stage4}) from Swin-B (using global attention for stage3 simultaneously) brings a performance improvement of $0.1\%$ which implies that the last stage is unnecessary. {\color{black}Replacing window attention with MLP blocks in the former 2 stages (\st{Win. Att.}) boosts the performance to $83.8\%$, which demonstrates that window attention is unnecessary in early stages.} The RPE is important and getting rid of that (\st{RPE}) will harm the performance about $0.3\%$. If we abandon the former 2 stages and down-sample $16\times$ using the patch embedding like plain ViT but increasing the block number to 24 (\st{Hierarchical}), the performance will decrease from $83.5\%$ to $82.9\%$. However, that is still higher than $81.8\%$ of plain ViT (\st{Deep}), which implies that hierarchical input module is important and a deeper architecture is much better than a shallow one.

    

\begin{table}[t]
\small
    \caption{Ablations of fully-supervised training on ImageNet-1K.}
    \vspace{-4pt}
 \begin{tabular}{lcc|cccc|cc|ccc}
    \toprule
    \multirow{2}{*}{Model}  & \multirow{2}{*}{Setting}  & \multirow{2}{*}{Dim}  & \multicolumn{4}{c|}{Depth}  & \multirow{2}{*}{RPE} & \multirow{2}{*}{\tabincell{l}{Win.\\Att.}} & Params  & FLOPs  & Top-1   \\
    &       &      & $56^2$   & $28^2$   & $14^2$   & $7^2$     &    &    & (M)     & (G)   & (\%)    \\
    \midrule
    Swin-B  &-
    & 512   & 2   & 2   & 18    & 2    & \checkmark   & \checkmark  & 88.0     & 15.4   & 83.5    \\
    
    \midrule
             & \st{Stage4}  & 512   & 2   & 2   & 20    & -    & \checkmark   & \checkmark  & 66.3   & 16.0   &
    83.6    \\
    HiViT-B  &\st{Win. Att.}    & 512   & 2   & 2   & 20    & -    & \checkmark   & \xmark  & 66.4   & 15.9   & \textbf{83.8}    \\
            & \st{RPE} & 512   & 2   & 2   & 20    & -    & \xmark    & \xmark  & 66.3   & 15.9   & 83.5    \\
            & \st{Hierarchical}         & 512   & -   & -   & 24    & -    & \xmark    & \xmark  & 76.7   & 15.8   & 82.9    \\
    ViT-B  &\st{Deep}
    & 768  & -  & -   & 12    & -    & \xmark    & \xmark  & 86.6   & 17.5   & 81.8    \\
    \bottomrule
    \label{tab.evolve}
\end{tabular}
\end{table}

\subsection{Self-Supervised Learning Results}

\paragraph{Experimental Details} For self-supervised pre-training, we use ImageNet-1K training dataset without using labels, and test the pre-trained models by fine-tuning and linear probing metrics in validation dataset. We inherit the pre-training settings of MAE~\cite{MAE2021} for pre-training. Specifically, we set the mask ratio to $75\%$ in default. The normalized target trick is also adopted. We use the AdamW optimizer~\cite{adamw} with the an initial learning rate of $1.5\times10^{-4}$, a weight decay of 0.05, and the learning rate follows the cosine decay learning schedule with a warm-up for $40$ epochs. The batch size is set to 4096 and the input size is $224\times224$. The overall pipeline is an encoder-decoder framework and the decoder is designed to have 6 transformer layers followed by a reshape operation to cast the feature to $3\times224\times224$. As for data augmentation, we only employ random cropping and random horizontal flip.
We test HiViT-B model in this part and the model is pre-trained for $300$ and $800$ epochs, which are then evaluated using fine-tuning and linear-probing metrics. As for fine-tuning, we inherit the training settings from ~\cite{MAE2021} and all the models are trained for $100$ epochs using AdamW optimizer with a warm-up for $5$ epochs, a weight decay of 0.05, and input size of $224\times224$. We use the layer-wise learning rate decay of 0.65. The initial learning rate is set to $5\times10^{-4}$ and batch size is set to 1024. As for linear probing, we train all the models for 100 epochs using LARS~\cite{huo2021large} optimizer with the batch size of 16,384 and learning rate of 0.1.

\paragraph{Fine-tuning} {\color{black}The fine-tuning and linear probing results are provided in Tab.~\ref{tab:ssl_ft_lin} and only the encoder part is used to test.} As shown in the Tab.~\ref{tab:ssl_ft_lin}, the HiViT-B (300e) version achieves $0.4\%$ and $0.2\%$ performance improvement than MAE models which are pre-trained for 1,600 epochs. Our longer training schedule version (800e) attains the dominant result of $84.2\%$, which outperforms MAE (1600e) by $0.6\%$ and SimMIM (Swin) by $0.2\%$. Compared to other methods including CAE, BEiT and iBOT, etc, HiViT-B model also shows superior results: $+1.0\%$ for BEiT, $+0.6\%$ for CAE, and $+0.2\%$ for MaskFeat.

\paragraph{Linear Probing} We evaluated the pre-trained models using linear probing metric, where all the parameters of the encoder are frozen except for a learnable classifier layer. From Tab.~\ref{tab:ssl_ft_lin}, we can see that HiViT-B model achieves good result of $71.3\%$, which is the best performance compared to all the MIM based methods. For example, the 800e model surpasses MAE (1600e) by $3.3\%$ with fewer pre-training epochs and the same training settings. The result also outperforms SimMIM (ViT-B) by $2.6\%$ and CAE (ViT-B) by $2.8\%$. 

\begin{table}[t]
    \centering
    \vspace{-4pt}
    \caption{Self-supervised learning results. The fine-tuning and linear probing results.}
    \begin{tabular}{ll|c|ccc|cc}
        \toprule
        Method     & Network  & Params  & Supervision    & Encoder   & Epochs    & FT (\%)      & LIN (\%)    \\  
        \midrule
        BEiT~\cite{bao2021beit}   
                     & ViT-B    &86   & DALLE           & 100\%     & 400       & 83.2         & -           \\
        CAE~\cite{chen2022context}      
                     & ViT-B      &86   & DALLE           & 100\%     & 800       & 83.6         & 68.3        \\
        MaskFeat~\cite{wei2021masked}
                     & ViT-B    &86     & HOG             & 100\%     & 800       & 84.0         & -           \\
        SimMIM~\cite{xie2021simmim}
                     & ViT-B    &86     & Pixel           & 100\%     & 800       & 83.8         & 68.7        \\
        SimMIM~\cite{xie2021simmim}
                     & Swin-B    &86    & Pixel           & 100\%     & 800       & 84.0         & -           \\
        MAE~\cite{MAE2021}
                     & ViT-B    &86     & Pixel           & 25\%      & 1600      & 83.6         & 68.0        \\ 
        \midrule
        Ours          & HiViT-B    &66   & Pixel          & 25\%      & 300       & 83.8         & -           \\
        Ours          & HiViT-B    &66   & Pixel          & 25\%      & 800       & \bf84.2         & \bf71.3        \\
        \bottomrule
    \end{tabular}
    \label{tab:ssl_ft_lin}
\end{table}


\begin{wraptable}{r}{0.46\textwidth}
\vspace{-18pt}
\setlength{\tabcolsep}{1.8mm}
\fontsize{9.5}{11}\selectfont
\caption{Pre-training efficiency comparison with different input sizes. }
\vspace{5pt}
\centering
\begin{tabular}{c|ccc} 
        \toprule
             \multirow{2}{*}{Input Size}        & ViT-B      & Swin-B    & HiViT-B \\
                                                & (MAE)      & (SimMIM)  & (Ours) \\
         \midrule
             $192\times192$    & 7.2      & 14.2     & 7.4   \\
             $224\times224$    & 9.5      & 18.4     & 9.7   \\
        \bottomrule
        \end{tabular}
        \label{tab:inference}
\label{tab:inference}
\vspace{-10pt}
\end{wraptable}

\paragraph{Training Efficiency} \textcolor{black}{HiViT only requires the active tokens as inputs so that our method enjoys the efficiency during the MIM pre-training.} As shown in Tab.~\ref{tab:inference}, we report the pre-training speed of MAE (ViT-B), SimMIM (Swin-B), and our HiViT-B with different input sizes. All the results represent the pre-training time (minutes) of 1 epoch on $8\times$ V100 GPUs. As the input image is $192\times192$, HiViT-B only takes 7.4 minutes per epoch, which is faster about $1.9\times$ than SimMIM and comparable with MAE. HiViT-B takes about 9.7 minutes when the input is $224\times224$, which is $1.9\times$ faster than SimMIM and comparable with MAE. We note that we use 6 decoder blocks with 512 dimension in default. Decreasing the decoder block number or the dimension also accelerate the pre-training speed without affecting our results. For example, setting the decoder block number to 4 and dimension to 384 only requires about 8 minutes per epoch and achieves $83.8\%$ performance.

\begin{wraptable}{r}{0.5\textwidth}
\vspace{-19pt}
\setlength{\tabcolsep}{1.8mm}
\fontsize{9.5}{11}\selectfont
\centering
\caption{Ablations by pre-training for 300 epochs and fine-tuning for 100 epochs (FT). The $56^2$, $28^2$, and $14^2$ denote the 1st, 2nd and 3rd stage respectively. The ID \textit{\#0} is the default setting.}
\vspace{5pt}
\begin{tabular}{c|ccc|cc|c} 
        \toprule
        \multirow{2}{*}{ID}   & \multicolumn{3}{c|}{Depth}  & Params  & FLOPs  & FT     \\
                              & $56^2$   & $28^2$  & $14^2$  & (M)    & (G)    & (\%)   \\
         \midrule
         \textit{\#0}   &2   &2   &20   &66.4   &15.9   &83.8 \\
         \#1   &1   &1   &22   &71.8   &15.9   &\textbf{83.9}  \\
         \#2   &2   &0   &22   &71.1   &15.9   &83.7  \\
         \#3   &0   &2   &22   &72.3   &15.9   &83.6  \\
         \#4   &0   &0   &24   &77.1   &16.0   &83.6  \\
        \bottomrule
        \end{tabular}
\label{tab:ablation}
\end{wraptable}

\paragraph{Ablation Studies} We perform some experiments to ablates our methods (see Tab.~\ref{tab:ablation}) and all the results are attained by pre-training for $300$ epochs. As shown, we test the self-supervised learning pre-training by setting different block number for stage 1, 2, and 3 (referred as to $56^2$, $28^2$, and $14^2$ respectively). The default setting (\textit{\#0}) achieves $83.8\%$ performance with $2-2-20$ block setting. Decreasing the block number for stage 1, 2 and increasing for stage 3 (\#1) brings more parameters and better performance of $83.9\%$, which is comparable to SimMIM result ($84.0\%$) by pre-training for $800$ epochs with Swin-B. We note that the $71.9$M parameters are still much lower than $87.8$M of Swin-B. Removing the stage-1 (\#3) or stage-2 (\#2) both harm the performance to $83.6\%$ and $83.7\%$ respectively, which demonstrates that the hierarchical architecture before the main stage is important and brings performance improvement. In addition, the result of \#3 is lower than \#2 denotes that the first stage seems more important than the second one. Removing both the former 2 stages attains a result of $83.6\%$, which further verifies the importance of the hierarchical architecture.

\subsection{Transfer to Dense Prediction Tasks}

\paragraph{Experimental Details} We transfer the self-supervised pre-trained models above to object detection on COCO and semantic segmentation on ADE20K. We follow the convention to perform object detection and semantic segmentation experiments. For the COCO experiments, we use the Mask R-CNN~\cite{MaskRCNN2017} head implemented by MMDetection library~\cite{MMdet2019}. We use the AdamW optimizer~\cite{adamw} with an initial learning rate of $3\times10^{-4}$ which decays by $10\times$ after the 9-th and 11-th epochs. The layer-wise decay rate is set to 0.75 and \textbf{$1\times$} training schedule (12 epochs) is adopted. We also apply multi-scale training strategy and single-scale testing. 
For ADE20K, we use the UperNet~\cite{xiao2018unified} head following BEiT~\cite{bao2021beit}. We also choose AdamW optimizer and the learning rate is $4\times10^{-4}$. We totally train the model for 160 iterations and the batch size is 16. The input resolution is $512\times512$ without using multi-scale testing. 

\vspace{-6pt}
\paragraph{Objection Detection on COCO.}
We transfer the same settings of CAE~\cite{chen2022context} to test our model in MS-COCO. We choose the 5-, 9-, 13-, 19-th blocks as inputs for later FPN network.
As shown in Tab.~\ref{tab:ssl_coco_ade}, we compare the performance with the state-of-the-art methods. Compared to BEiT~\cite{bao2021beit} (we bollow the results from CAE~\cite{chen2022context}), HiViT-B shows superior results over $7.4\%$ AP$^{\text{box}}$ and $6.0\%$ AP$^{\text{mask}}$ respectively. 
MAE~\cite{MAE2021} (1600e) achieves the $48.4\%$ AP$^{\text{box}}$ and $42.6\%$ AP$^{\text{mask}}$ results, which is lower than our $49.5\%$ and $43.8\%$ respectively.
CAE~\cite{chen2022context} improves the performances to $49.2\%$ AP$^{\text{box}}$ and $43.3\%$ AP$^{\text{mask}}$ by pre-training for 800 epochs. But the results still are below than our results by $0.3\%$ and $0.8\%$ respectively even though CAE~\cite{chen2022context} uses image tokenzier described in DALLE~\cite{ramesh2021zero}.

\vspace{-6pt}
\paragraph{Semantic Segmentation on ADE20K.}
The results on ADE20K are shown in Tab.~\ref{tab:ssl_coco_ade}. We note that we do not introduce any other tricks and test all the models using the same settings. 
We report the mean intersection over union (mIoU) performances. As shown, MoCo-v3 reports the $47.3\%$ mIoU result by pre-training for $300$ epochs, which is lower than our $48.3\%$. BEiT~\cite{bao2021beit}, CAE\cite{chen2022context}, and MAE~\cite{MAE2021} report the performance of $47.1\%$, $48.8\%$, and $48.1\%$ respectively. By pre-training for 1600 epochs, MAE achieves the $48.1\%$ mIoU. Compared to these state-of-the-art methods, HiViT-B, by pre-training for 800 epochs, reports the $48.3\%$ result, which is higher than all the methods in apart from CAE, which uses tokenizer of DALLE~\cite{ramesh2021zero}.

\begin{table}[]
    \centering
    \caption{Downstream task fine-tuning results transferred from self-supervised pre-training.}
    \begin{tabular}{l|c|c|l|cc|c}
		\toprule
		\multirow{2}{*}{Method} &\multirow{2}{*}{Network}  &\multirow{2}{*}{Params}  &\multirow{2}{*}{Pre-train data}   & \multicolumn{2}{c|}{COCO} & ADE20K \\
{}&  &  & {} & AP$^{\text{box}}$ & AP$^{\text{mask}}$ & mIoU \\ 
		\midrule
		Supervised~\cite{MAE2021} & ViT-B  &86  & IN1K w/ labels  & 47.9 & 42.9 & 47.0 \\
		MoCo v3~\cite{chen2021empirical} & ViT-B  &86  & IN1K & 45.5  & 40.5  &47.3\\
		BEiT~\cite{bao2021beit} & ViT-B  &86 & IN1K+DALLE & 42.1  & 37.8   & 47.1\\
		CAE~\cite{chen2022context}  & ViT-B   &86  & IN1K+DALLE & 49.2  & 43.3  & \bf48.8 \\
		MAE~\cite{MAE2021} (1600e) & ViT-B  &86  & IN1K  & 48.4 & 42.6  &48.1\\
		\midrule
		Ours (800e) & HiViT-B  &66  & IN1K  &\bf49.5  &\bf43.8  & 48.3\\
		\midrule
	\end{tabular}
    \label{tab:ssl_coco_ade}
\end{table}


\section{Conclusions}

This paper presents a hierarchical vision transformer named HiViT. Starting with Swin Transformers, we remove redundant operations that cross the border of tokens in the main stage, and show that such modifications do not harm, but slightly improve the model's performance in both fully-supervised and self-supervised visual representation learning. HiViT shows a clear advantage in integrating with masked image modeling, on which the efficient implementation on ViT can be directly transplanted, accelerating the training speed by almost $100\%$. We expect that HiViT becomes an off-the-shelf replacement of ViT and Swin Transformers in the future research.

\vspace{-6pt}
\paragraph{Limitations} Despite the improvement observed in the experiments, our method has some limitations. The most important one lies in that the masking unit size is fixed -- this implies that we need to choose a single `main stage'. Fortunately, the 3rd stage of Swin Transformers contribute most parameters and computations, hence it is naturally chosen, however, the method may encounter difficulties in the scenarios that no dominant stages exist. In addition, we look forward to more flexible architecture designs that go beyond the constraints -- a possible solution lies in modifying low-level code (\textit{e.g.}, CUDA) to support arbitrary and variable grouping of tokens.

\vspace{-6pt}
\paragraph{Societal Impacts} Our research focus on (i) designing efficient architectures for deep neural networks and (ii) self-supervised learning. These two topics have been widely studied in the community, and our work does not have further societal impacts than others.


\section*{\Large Appendix}
\appendix

\section{Architecture Comparison of Fully-Supervised Learning on ImageNet-1K}
We compare different transformer architectures including \textbf{plain transformers}, \textbf{hierarchical transformers}, and \textbf{hybrid ones} on fully-supervised learning using ImageNet-1K and the results are shown in Tab.~\ref{tab:image_classification}. We can see that plain transformers usually do not contains local inter-unit operations, which implies that these models are structurally simple generally. However, these models suffer poor performance than hierarchical ones. To capture hierarchical property for transformers, researcher introduce complex local inter-unit operations (such as spatial-reduction attention and window attention, etc, as shown in the table) to plain transformer, so that the models cater to visual priors and attain better results. Our HiViT, as shown, enjoys the hierarchical attributes without local inter-unit operations, which makes our model structurally simple and MIM friendly. In addition, HiViT gets better results without bells and whistles. The hybrid transformers bring convolutional layers (expect for patch embedding) to transformer and usually attain the dominant results than the former two. But these models are unfriendly to MIM self-supervised task because of the sliding prior of convolutional layers. Moerover, our self-supervised HiViT-B achieves competitive result of $84.2\%$.

\setcounter{table}{7}
\begin{table}[h]
    \setlength{\tabcolsep}{0.13cm}
    \centering
    \caption{ImageNet-1K results of different transformer architectures.}
    \begin{tabular}{ll|l|ccc}
    \toprule
    \multirow{2}{*}{Model type}                                                & \multirow{2}{*}{Model}    
                                & \multirow{2}{*}{Local inter-unit operations} & Params      & FLOPs       & Top-1   \\
    &                           &                                              & (M)         & (G)         & (\%)    \\
    \midrule
    \multirow{5}{*}{\tabincell{c}{Plain \\ Transformers}}                        & ViT-B/16~\cite{DeiT2021} 
                                & -                                            & 86.7        & 17.4        & 81.8    \\
                                                                               & CrossViT-B~\cite{chen2021crossvit}
                                & -                                            & 104.7       & 21.2        & 82.2    \\
                                                                               & CaiT-M24~\cite{touvron2021going} 
                                & -                                            & 185.9       & 36.0        & 83.4    \\
                                                                               & T2T-ViT-24~\cite{yuan2021tokens}
                                & -                                            & 64.1        & 14.1        & 82.3    \\
                                                                               & TNT-B~\cite{han2021transformer}
                                & -                                            & 65.6        & 14.1        & 82.8    \\
    \midrule
    \multirow{4}{*}{\tabincell{c}{Hierarchical\\Transformers}}                 & PVT-L~\cite{wang2021pvtv2}
                                & spatial-reduction attention                  & 61.4        & 9.8         & 81.7    \\
                                                                               & ViL-B~\cite{zhang2021multi}
                                & window attention                             & 55.7        & 13.4        & 83.2    \\
                                                                               & Swin-B~\cite{liu2021swin}  
                                & shifted window attention                     & 87.8        & 15.4        & 83.5    \\
                                                                               & HiViT-B (Ours) 
                                & -                                            & 66.4        & 15.9        & \bf83.8 \\
    \midrule
    \multirow{4}{*}{\tabincell{c}{Convolution \\ $+$ \\ Transformers}}                
                                                                               & Conformer\cite{Conformer2021}     
                                & CNN branch                                   & 83.3        & 23.3        & 84.1    \\
                                                                               & CoAtNet-2\cite{dai2021coatnet} 
                                & depth-wise convolution                       & 75.0        & 15.7        & 84.1    \\
                                                                               & CSwin-B\cite{dong2021cswin} 
                                & $\left\{\begin{array}{l} \text{convolution in LePE}  \\ \text{cross-shaped window attention} \end{array} \right.$
                                                                               & 78.0        & 15.0        & 84.2    \\
    \midrule
    Self-supervised                                                            & HiViT-B (Ours) 
                                & -                                            & 66.4        & 15.9        & \bf84.2 \\
    \bottomrule
    \end{tabular}
    \label{tab:image_classification}
\end{table}

\section{Improved Results on COCO and ADE20K}

In dense prediction tasks such as object detection and semantic segmentation, the feature pyramid is a key component.
Our experiments in Tab.~\ref{tab:ssl_coco_ade} use the same settings as the ViT model using self-supervised pre-training in ~\cite{MAE2021, chen2022context}. 
This setting extracts intermediate features and up-samples/down-samples them by deconvolution/convolution for pyramid feature generation, which works for plain transformers but does not take advantage of our hierarchical structure.

In order to fully exploit the capabilities of our hierarchical vision transformer, we perform an improved experimental setting following~\cite{liu2021swin}. We use the three resolution features (with strides of 4, 8, 16) generated by stage-1/-2/-3, and add a stride-32 feature which is down-sampled from the stage-3 block to align with the standard feature pyramid generation. 
With this change, HiViT achieves improved results on COCO and ADE20K, as shown in Tab.~\ref{tab:dense}. In particular, it achieves 51.2\% AP$^{\text{box}}$, 44.2\% AP$^{\text{mask}}$ in COCO and 51.2\% mIoU in ADE20K.

\begin{table}[h]
    \centering
    \caption{Improved downstream task fine-tuning results on COCO and ADE20K.}
    \begin{tabular}{l|c|c|l|cc|c}
		\toprule
		\multirow{2}{*}{Method}           & \multirow{2}{*}{Network}                 & \multirow{2}{*}{Params}     & 
		\multirow{2}{*}{Pre-train data}   & \multicolumn{2}{c|}{COCO}                & ADE20K                      \\
                                          &                                          &                             & 
                                          & AP$^{\text{box}}$   & AP$^{\text{mask}}$ & mIoU                        \\ 
		\midrule
		Supervised~\cite{MAE2021}         & ViT-B                                    & 86                          & 
		IN1K w/ labels                    & 47.9                & 42.9               & 47.0                        \\
		MoCo v3~\cite{chen2021empirical}  & ViT-B                                    & 86                          & 
		IN1K                              & 45.5                & 40.5               & 47.3                        \\
		BEiT~\cite{bao2021beit}           & ViT-B                                    & 86                          & 
		IN1K+DALLE                        & 42.1                & 37.8               & 47.1                        \\
		CAE~\cite{chen2022context}        & ViT-B                                    & 86                          & 
		IN1K+DALLE                        & 49.2                & 43.3               & 48.8                        \\
		MAE~\cite{MAE2021} (1600e)        & ViT-B                                    & 86                          & 
		IN1K                              & 48.4                & 42.6               & 48.1                        \\
		\midrule
		Ours                              & HiViT-B                                  & 66                          & 
		IN1K                              & 49.5                & 43.8               & 48.3                       \\
		Ours (improved)                   & HiViT-B                                  & 66                          & 
		IN1K                              & 51.2                & 44.2               & 51.2                        \\
		\midrule
	\end{tabular}
    \label{tab:dense}
\end{table}

\section{The Processing Pipeline of HiViT in MIM Pre-training}

In MIM pre-training, HiViT uses the encoder to process the visible tokens of the input image as shown in Fig.~\ref{fig:pipeline}. Here, since we have multi-resolution image patches, we call the smallest unit that may be masked as a mask-unit. For an input image, we first align the patch embedding to get a feature vector of shape $M\times$4$\times$4$\times$128, where $M$ is the total number of all mask-units (\textit{e.g.}, $M=196$ according to the common practice). Then, we randomly select a certain proportion ($25\%$) of mask-units to obtain a feature vector of $M'\times$4$\times$4$\times$128, where $M'$ is the number of visible mask-units which is often much smaller than $M$. HiViT will process the features of the visible mask-units stage by stage, and finally get $M'\times$512 vectors as the output of encoder and feed them to the decoder for pixel restoration. 

During the pre-training process, all three stages only need to process visible mask-units -- this is why our model preserves the hierarchical structure and has efficient self-supervised pre-training efficiency. 

\setcounter{figure}{2}
\begin{figure}[h]
    \centering
    \includegraphics[width=1.0\linewidth]{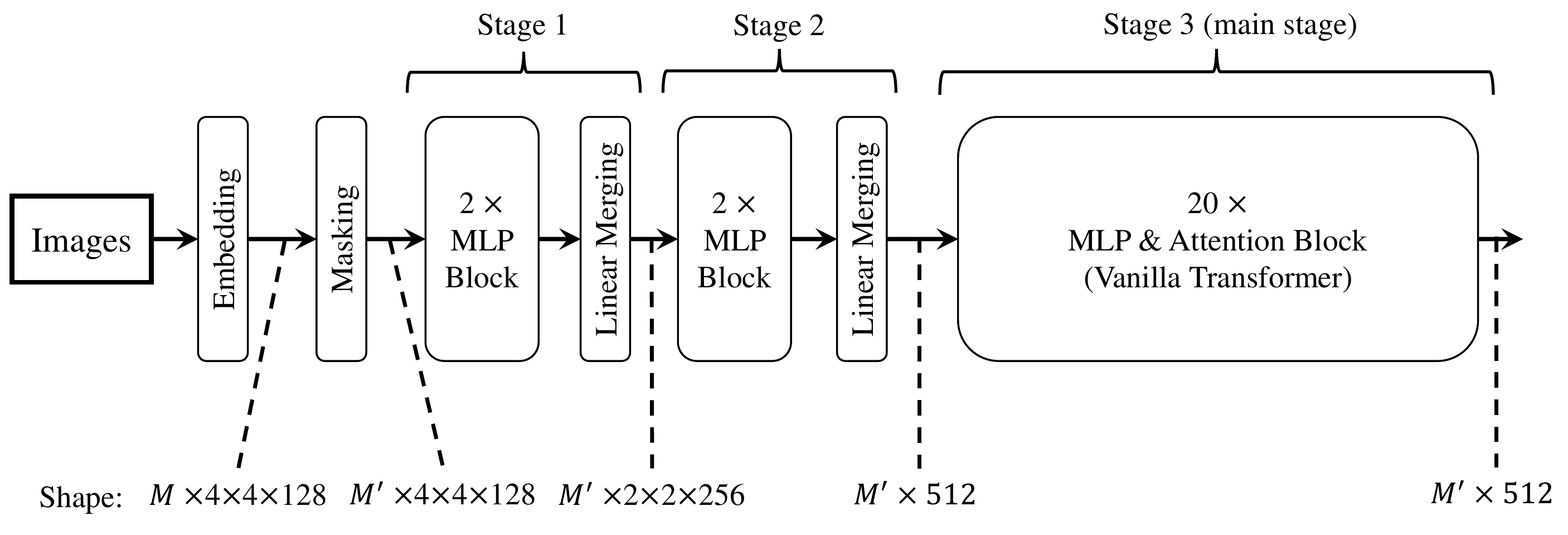}
    \caption{Pipeline of HiViT in MIM pre-training.}
    \label{fig:pipeline}
\end{figure}

{
\small
\bibliographystyle{plain}
\bibliography{output.bbl}
}

\end{document}